\pdfoutput=1

\documentclass[11pt]{article}

\usepackage{acl}

\usepackage{times}
\usepackage{latexsym}
\usepackage{makecell}
\usepackage[T1]{fontenc}

\usepackage[utf8]{inputenc}

\usepackage{microtype}

\usepackage{inconsolata}

\usepackage{graphicx}
\usepackage{subcaption}
\usepackage{colortbl}
\usepackage{rotating}
\usepackage{multirow}
\usepackage{amsmath}
\usepackage{arydshln}
\usepackage{amsfonts}
\usepackage{dcolumn} 
\usepackage{xspace}
\usepackage{svg}
\usepackage{algorithmic}
\usepackage{algorithm}
\usepackage{amsmath}
\usepackage{amssymb}
\usepackage{mathtools}
\usepackage{amsthm}
\usepackage{booktabs}

\title{Enhancing Cross-Tokenizer Knowledge Distillation with \\ Contextual Dynamical Mapping}

\author{Yijie Chen$^1$, Yijin Liu$^2$, Fandong Meng$^2$, Yufeng Chen$^1$, Jinan Xu$^1$, Jie Zhou$^2$ \\
  $^1$Beijing Key Lab of Traffic Data Analysis and Mining, \\
Beijing Jiaotong University, Beijing, China \\
  $^2$Pattern Recognition Center, WeChat AI, Tencent Inc, China \\
    {\tt \{22120354, chenyf, jaxu\}@bjtu.edu.cn}\\
    {\tt \{yijinliu, fandongmeng, withtomzhou\}@tencent.com}
 }

\begin{document}
\maketitle
\begin{abstract}

Knowledge Distillation (KD) has emerged as a prominent technique for model compression. However, conventional KD approaches primarily focus on homogeneous architectures with identical tokenizers, constraining their applicability in cross-architecture scenarios. As for the cross-tokenizer KD, the differences in the tokenizers give rise to two fundamental challenges: (1) sequence misalignment caused by divergent tokenization strategies, and (2) mismatched vocabulary size and composition.
While existing probability-matching methods attempt to address these issues, their efficacy remains limited due to suboptimal alignment in both the sequence and vocabulary aspects. To overcome these limitations, we propose Contextual Dynamic Mapping (CDM), a novel cross-tokenizer distillation framework that employs contextual information to enhance sequence alignment precision and dynamically improves vocabulary mapping.
We evaluated the effectiveness of our approach across five advanced and widely-used model families ({\em i.e, }LLama3, Phi3, Gemma2, OPT and Qwen2), which were configured into three distinct teacher-student pairs. Our method shows significant advantages over existing cross-tokenizer distillation baselines across diverse benchmarks, including instruction-following, code generation and math. Notably, our analysis reveals that combining conventional same-tokenizer distillation and cross-tokenizer distillation through CDM yields further performance improvements. The code is  available at \url{https://github.com/pppa2019/ContexualDynamicMapping}

\end{abstract}

\begin{figure}[ht]
  \centering
  \begin{subfigure}[b]{\linewidth}
    \centering
    \includegraphics[width=0.9\textwidth]{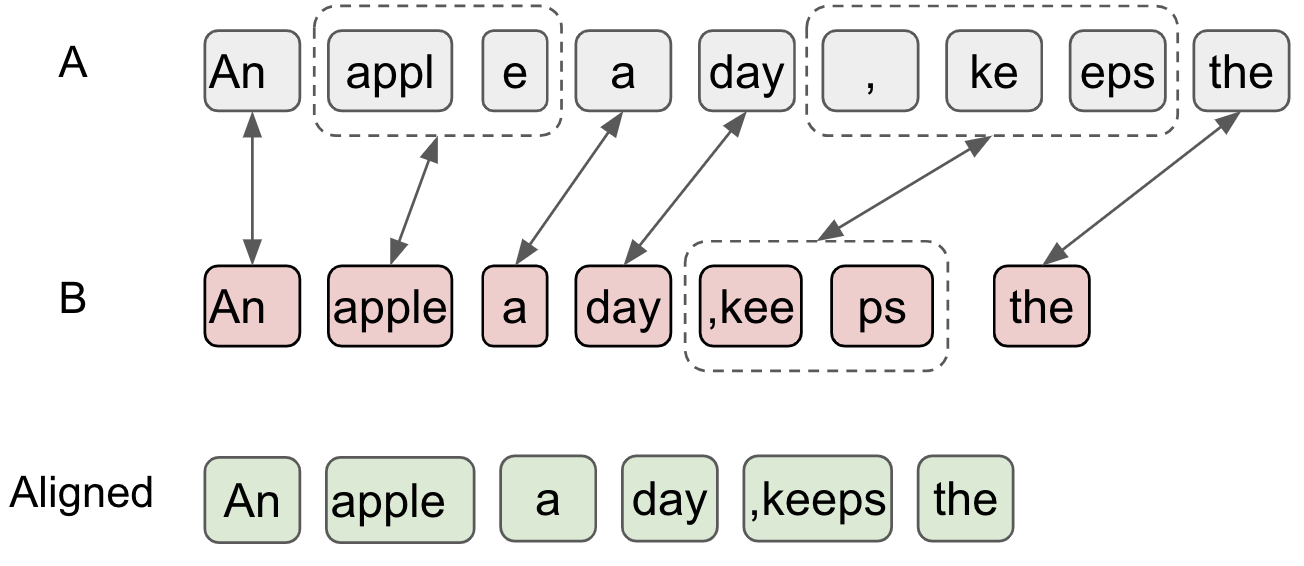}
    \caption{First, the token sequences of teacher and student models are aligned to spans consisting of similar text.}
  \end{subfigure}
  ~
  \begin{subfigure}[b]{\linewidth}
    \centering
    \includegraphics[width=0.9\textwidth]{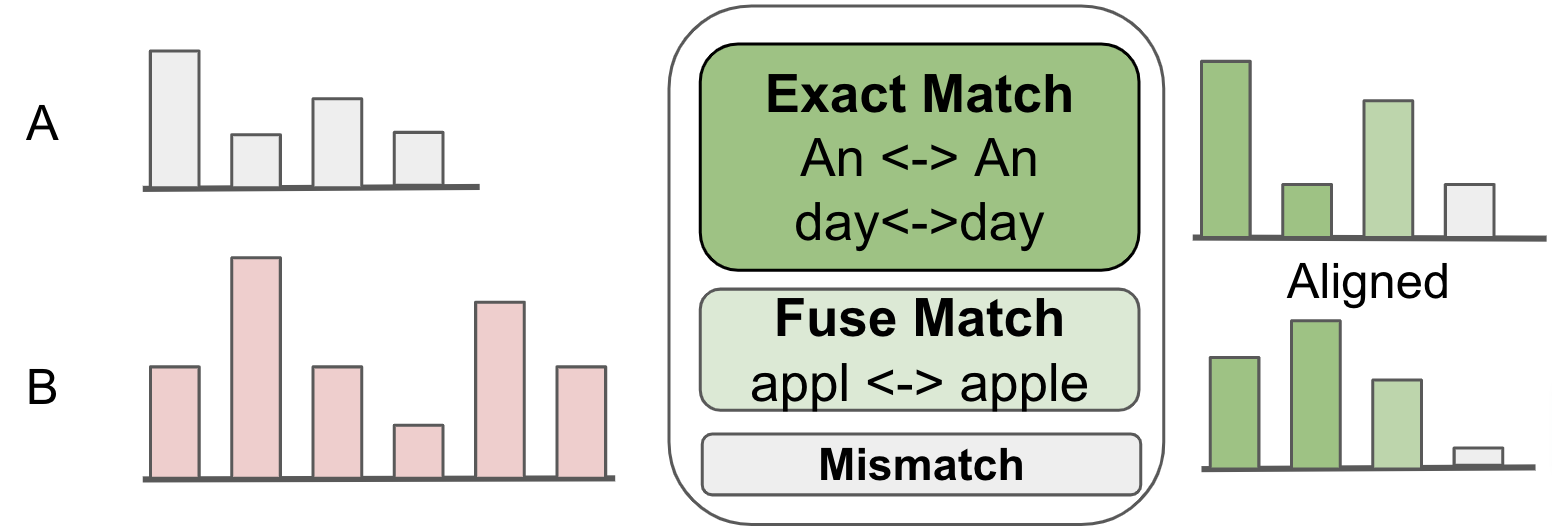}
    \caption{The vocabulary distribution from the teacher and student models should be aligned via token mapping.}
  \end{subfigure}
  \caption{The illustration of the alignment process of cross-tokenizer knowledge distillation. A and B mean the tokenizers of the student or teacher models.} 
  \label{fig:task}
\end{figure}

\section{Introduction}
Knowledge distillation (KD)~\cite{hinton2015distilling, wen2023f, gu2024minillm, ko2024distillm, guo2025deepseek} has emerged as a promising methodological framework for enhancing the performance of compact models through knowledge transfer from larger, more powerful teacher models.
Nevertheless, conventional KD approaches, which aim to minimize the distribution difference between the logits of teachers and students, are still restricted by the requirement for tokenizer consistency between teachers and students.
To address this limitation, Cross-Tokenizer KD (CTKD)~\cite{fu2023specializing,boizard2024towards} has emerged as a critical research frontier.
As illustrated in Figure~\ref{fig:task}, the core challenges of CTKD arise from two fundamental aspects: (1) divergent tokenization strategies induce sequence misalignment during text processing, and (2) vocabulary discrepancies create dimensional and semantic mismatches in output probability spaces.
Both the sequence and vocabulary level misalignments create significant barriers to effective knowledge transfer between heterogeneous architectures.
Current approaches attempt to bridge these gaps through two primary strategies:
\begin{itemize} 
\item Tokens mapping based on text character similarity~\cite{fu2023specializing,wanknowledge,wan2024fusechat}, which risks semantic misalignment ({\em e.g.,} confusing "denoted" and "devoted") 
\item Optimal transport methods~\cite{boizard2024towards,cui2024multi} that compute full distribution distances but lack explicit semantic alignment. 
\end{itemize}

To quantify the discrepancies in sentence tokenization and vocabulary among mainstream large language models (LLMs) equipped with different tokenizers, we conducted a comprehensive analysis of alignment rates\footnote{The detail statistic process is shown at Section~\ref{sec:stat}} across five mainstream LLMs. The results demonstrate the wide range of sequence alignment rates (around 30\%–90\%) and vocabulary alignment rates (around 10\%–90\%), which collectively highlight the substantial room for improvement in cross-tokenizer knowledge transfer.
To further enhance the accuracy of model output alignment, we introduce Contextual Dynamic Mapping (CDM), a novel CTKD framework that introduces: \begin{itemize} \item \textbf{Sequence-level}: Entropy-weighted dynamic programming, which improves the precision of sequential token alignment by dynamically adjusting the alignment process utilizing entropy-based measure of tokens. 
\item \textbf{Vocabulary-level}: Context-aware candidate matching dynamically constructs semantic-preserving token mappings, achieving an improved balance between token-level precision and minimization of unmapped token pairs.
\end{itemize}
Building on this mapping framework, we futher prioritize Top-K tokens based on their contextual significance, which effectively suppresses noise while enhancing reasonable token mapping rates further.

The experiments are based on five encompassing open-source model series, and the training and evaluation contain various tasks, including instruction following, code generation, and math.
The experimental results demonstrate that CDM exhibits consistent and substantial superiority over current mainstream cross-tokenizer distillation approaches among different tasks ({\em e.g.,} for Qwen2-0.5B model, CDM has the improvement of 4.27\% on instruction following tasks, 12.19\% on code tasks and 3.34\% on math tasks) and even surpasses the performance of same-tokenizer distillation in some settings. Our main contributions are as follows:
\begin{itemize}
    \item We propose the CDM method, which facilitates cross-tokenizer distillation through contextual information to improve the matching accuracy in sequence and vocabulary aspects.
    \item Extensive experiments are conducted on various backbone models and datasets across three tasks, and the experimental results indicate the effectiveness of CDM consistently.
    \item We provide a detailed analysis of the alignment rate improvement in CDM to illustrate the correlation between alignment and distillation effectiveness.
    \item We observe that the model performance can be further improved by combining the conventional same-tokenizer distillation and CDM.
\end{itemize}

\begin{figure}
    \centering
    \includegraphics[width=0.7\linewidth]{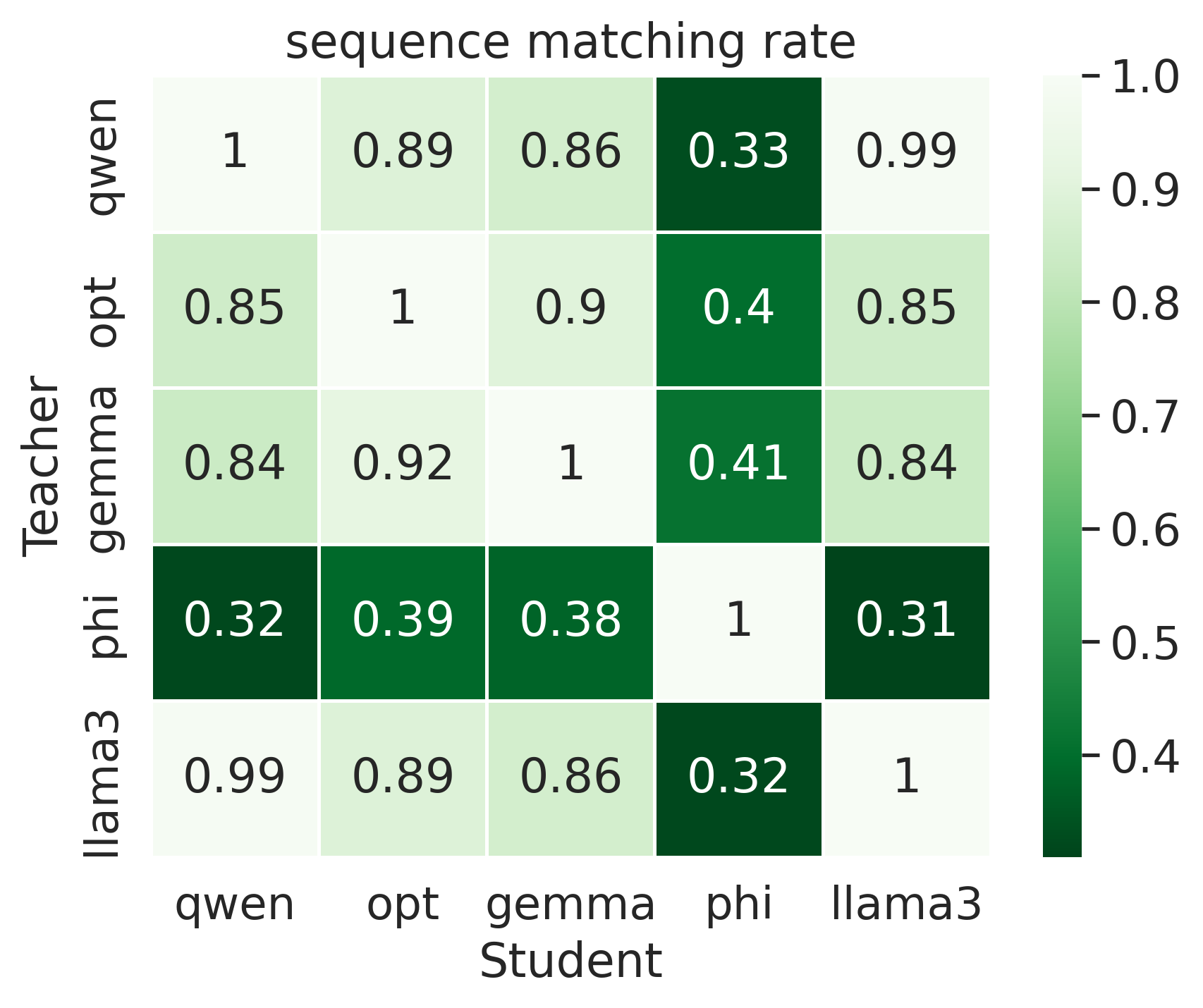}
    \caption{Matching rate of sequence alignment results.}
    \label{fig:seq_rate}
\end{figure}
\begin{figure}
    \centering
    \includegraphics[width=0.7\linewidth]{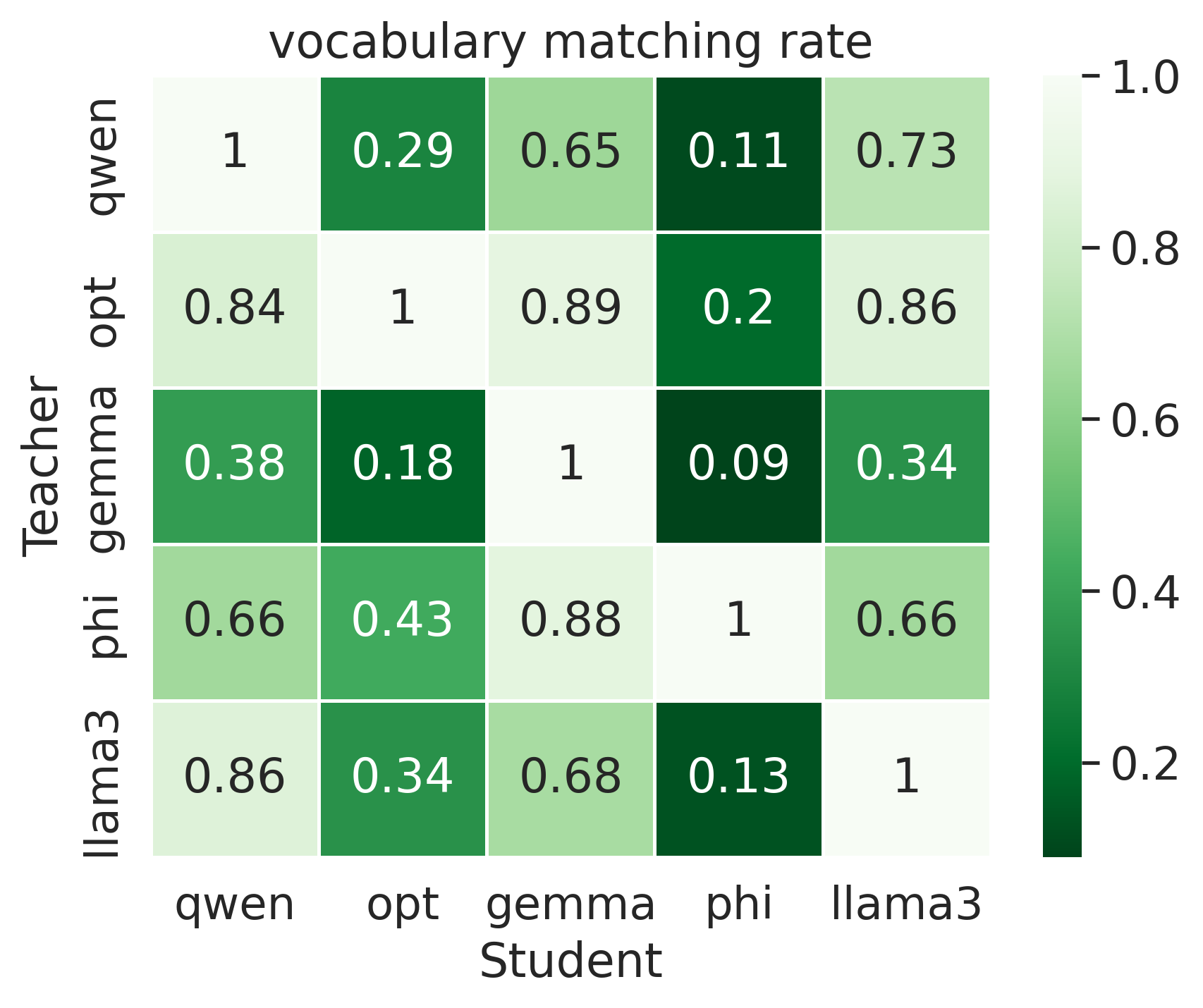}
    \caption{Matching rate of vocabulary alignment results.}
    \label{fig:vocab_rate}
\end{figure}
\begin{figure*}
    \centering
    \includegraphics[width=0.95\linewidth]{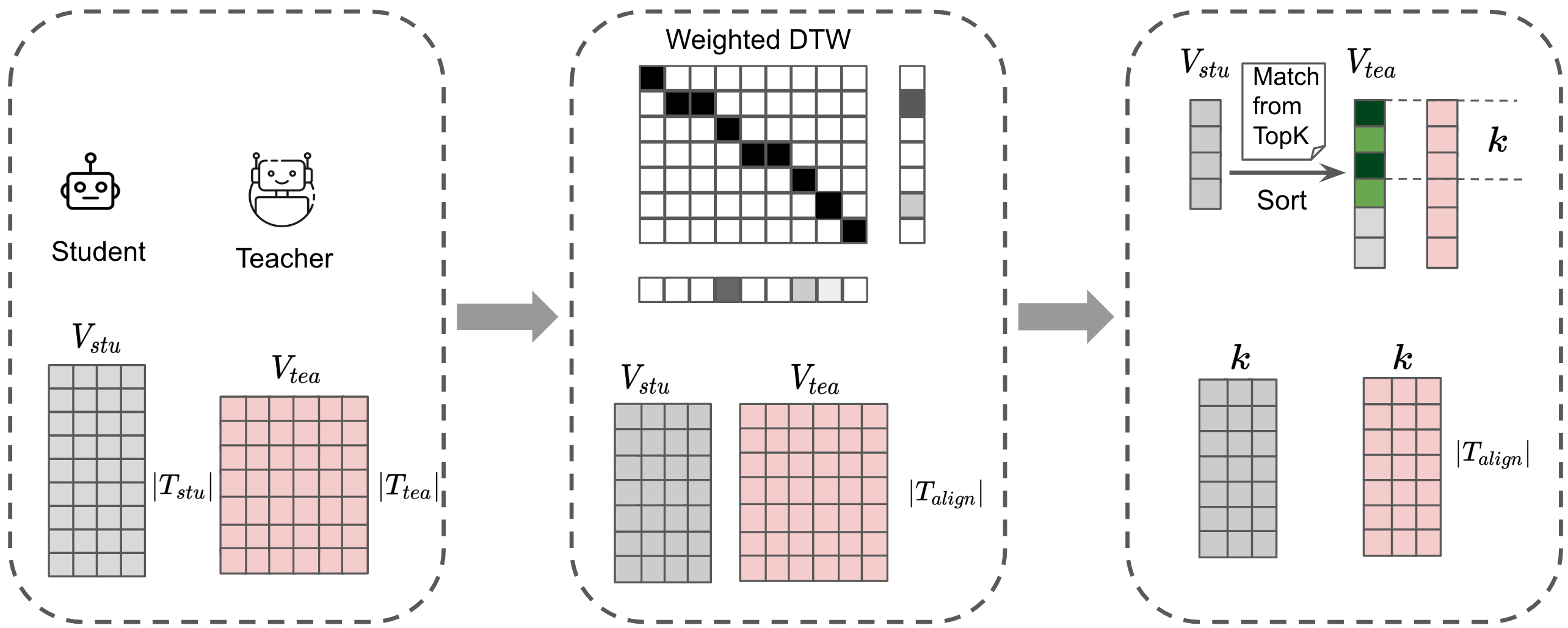}
    \caption{The architecture of CDM consists of two key components: an entropy-weighted Dynamic Time Warping (DTW) sequence alignment algorithm and a dynamic Top-K vocabulary mapping algorithm. Following the mapping procedure, the output representations from both the teacher and student models are aligned to ensure consistency in both dimensional structure and semantic space.}
    \label{fig:CDM}
\end{figure*}
\section{Preliminary Analysis}\label{sec:stat}
To systematically investigate the fundamental challenges in CTKD, we analyze the alignment rates in sequence and vocabulary level separately.
On the sequence level, we calculate the tokenization result of the two selected tokenizers. Concretely, for a certain sentence, we statistic the set of tokens overlap and average on the sample number as final results. The statistic is based on 3000 examples sampled from the training corpus of Dolly-15K~\cite{ouyang2022training}
On the vocabulary level, we calculate the ratio of tokens that exactly matches the two selected tokenizers.

The statistic results presented in Figure~\ref{fig:seq_rate} and Figure~\ref{fig:vocab_rate} highlight two critical limitations: (1) vocabulary-level alignment rates, which reach as low as 9\%, and (2) sequence-level alignment rates, with minimum values as low as 31\%. 
These findings underscore the critical need for supplementary mapping mechanisms within existing methodologies. Furthermore, the intersection of these two mismatch dimensions creates compounding effects, namely, sequence alignment errors amplify vocabulary mismatches through error propagation.
This issue motivates our proposed contextual dynamic mapping framework, which mitigates alignment errors by utilizing contextual information.

\section{Contextual Dynamic Mapping}
\subsection{Formal Definition}
In this section, we take subscripts $stu$ and $tea$ to denote the student and teacher models, respectively. 
Given an input sentence, the tokenization process yields token sequences $T_{stu}$ and $T_{tea}$, along with the vocabulary size $|V_{stu}|$ and $|V_{tea}|$ ($|V_{stu}|\neq|V_{tea}|$ and $ |T_{stu}|\neq |T_{tea}|$). The model output logits are then expressed as $O_{stu} \in \mathbb{R}^{|V_{stu}|\times|T_{stu}|}$ and $O_{tea}\in \mathbb{R}^{|V_{tea}|\times|T_{tea}|}$.

\subsection{Sequence Alignment with Contextual Weight} 
\citet{fu2023specializing} introduced a method based on Dynamic Time Warping (DTW) for sequence token alignment, later adopted in~\cite{wanknowledge,wan2024fusechat}. The DTW is a dynamic programming method that can minimize the cost $D_{\text{DTW}}$ between the two token lists for the same sentence tokenized by different tokenizers. The algorithms minimize:
\begin{equation}
    D_{\text{DTW}}(T_{stu}, T_{tea}) = \min_{\pi \in \Pi} \sum_{(i,j) \in \pi} cost(t_i^{stu}, t_j^{tea})
\end{equation}
where $\Pi$ is the set of alignment paths. In existing works~\cite{fu2023specializing,wan2024fusechat}, the cost function of DTW $cost(\cdot)$ use edit distance $EditDistance(t_i^{stu}, t_j^{tea})$.
However, this approach demonstrates suboptimal performance with occasionally generating non-sensical alignments, for the edit distance metric introducing misalignment. 

To overcome these constraints, we propose an entropy-enhanced DTW to incorporate the context information into the alignment process. 
The entropy-weight prioritizes the more informative tokens while suppressing noisy alignments, mitigating the sequence token mismatch problem. 
To illustrate, consider the example in Figure~\ref{fig:task}. For the original DTW, the edit distance between \textit{``e''} (from A, denote with \textit{italics}) and ``apple'' (from B) is larger than the edit distance between \textit{``e''} and ``a''(from B), which leads to the mistaken matching between the \textit{``e''} and ``a''. 
However, the entropy in ``a'' and \textit{``e''} both have higher entropy compared with ``apple'' for their ambiguity, so after adding entropy-based weight the distance metric between \textit{``e''} and ``apple'' will be lower, leading to a correct mapping (\textit{``appl''}, ``apple'') and (\textit{``e''}, ``apple'').

For a tokenizer with vocabulary size $V$, let $m$ denote any model in which both can be a teacher or student, $O_m \in \mathbb{R}^{|T_m| \times |V_m|}$ represent the output logits where $|T_m|$ denotes the length of the token sequence and $o_i \in R^{|V_m|}$ presents the logits vector of the $i\text{-th}$ vector in $O_m$. We first compute position-wise entropy $H \in \mathbb{R}^{|T_m|}$ through:
\begin{equation}\label{eq:entropy}
\mathrm{H}(o_i) = -\sum_{j=1}^{|V_m|} p(o_i^j) \log p(o_i^j)
\end{equation} 
Subsequently, we apply the MinMax as the normalization function $\phi(X) = \frac{x - min(x)}{max(x)-min(x)}$ and conduct a linear mapping to obtain alignment weights.
\begin{equation}\label{eq:weight}
\mathrm{W}(O) = \lceil \mathrm{Sigmoid}(\phi(H(O))) \cdot C + C \rceil
\end{equation}
The hyperparameter $C$ controls the weighting range $[C, 2C]$ to ensure both flexibility and computational efficiency in the weighting process. After calculate the weight as Equation~\ref{eq:weight}, let $cost(t_i^{stu}, t_j^{tea} ) = w_i^{stu} \cdot w_j^{tea} \cdot EditDistance(t_i^{stu}, t_j^{tea})$ be the cost function of DTW.
With weighted DTW, we obtain span-level token mapping lists $T_{stu}^{seq}$ and $T_{tea}^{seq}$. The original logits $O_{tea}$ and $O_{stu}$ are merged according to these mapping lists using mean pooling. After that, the outputs $O^{seq}_{stu} \in \mathbb{R}^{|T_{align}|\times |V_{stu}|}$ and $O^{seq}_{tea} \in \mathbb{R}^{|T_{align}|\times |V_{tea}|}$ are obtained and have been aligned in the sequence-level.

\begin{algorithm}[!h]
    \caption{Algorithm of token alignment}
    \label{alg:AOS}
    \renewcommand{\algorithmicrequire}{\textbf{Input:}}
    \renewcommand{\algorithmicensure}{\textbf{Output:}}
    
    \begin{algorithmic}[1]
        \REQUIRE $O_{stu}^{seq}$, $O_{tea}^{seq}$, $F_{EM}$, $\theta$
        \ENSURE $O_{stu}^{align}, O_{tea}^{align}$
        \STATE initialize $F_{dynamic} := F_{EM}$
        \STATE gets the tokens $T_{stu}^{topk}, T_{tea}^{topk}$ with Top-K logits in each position utilizing.
        \FOR { each position $i \in [1, |T_{align}|)$}
            \FOR { each $tok_a \in T_{stu}^{topk}[i]$}
                \IF {$tok_a \notin F_{EM}$}
                    \STATE $ best=\emptyset; min\_dist = \infty$
                    \FOR {each $tok_{b} \in T_{tea}^{topk}[i]$ }
                        \STATE $d = dist\_func(tok_a, tok_b)$
                        \IF {$d<\theta\ \textbf{\ and\ } d<min\_dist$}
                            \STATE $best=tok_a, min\_dist=d$ 
                        \ENDIF
                    \ENDFOR
                    \IF {$best \neq \emptyset$}
                        \STATE $F_{dynamic} := F_{dynamic} \cup \{tok_{b} \rightarrow best\}$
                    \ENDIF
                \ENDIF
            \ENDFOR
        \ENDFOR
        \STATE $O_{stu}^{align} = \text{Mask}(F_{dynamic}(O^{topk}_{tea}))$
        \STATE $O_{tea}^{align}=\text{Mask} (O^{topk}_{tea})$
    \end{algorithmic}
    \textbf{Return:} $O_{stu}^{align}, O_{tea}^{align}$
\end{algorithm}

\subsection{Dynamic Vocabulary Mapping with Contextual Candidates}\label{sec:cdm_voc}

After the sequence alignment, the model logits are still not aligned on the vocabulary dimension, {\em i.e.,} $|V_{stu}|\neq|V_{tea}|$. CDM uses contextual information to improve the vocabulary mapping accuracy, and the core process and is described as Algorithm~\ref{alg:AOS}.
First, tokens that can be exactly matched between the two tokenizers are stored in a mapping dictionary $F_{EM}$ to facilitate efficient mapping operations. For tokens that remain unmapped, we introduce a dynamic mapping dictionary $F_{dynamic}$, which is initialized as a copy of $F_{EM}$ (line 1). The process involves utilizing the aligned model logits $O^{seq}_{tea}$ and $O^{seq}_{stu}$. In order to preserve the most relevant information in the context for the training instance and avoid mapping between irrelevant tokens, we select the top $k$ logits at each position according to Equation~\ref{eq:topk}.
\begin{equation}\label{eq:topk}
\mathrm{Top}\text{-}K (o_i)=\operatorname{argsort}\left(\sum_{j=1}^{|V|} o_i^j, \downarrow\right)[:k]
\end{equation}

Then, $\mathrm{Top}\text{-}K(O_{stu}^{seq})$, $ \mathrm{Top}\text{-}K(O_{tea}^{seq})$ yields both the corresponding logits value $O_{tea}^{topk}\in\mathbb{R}^{|T_{align}|\times k}, O_{stu}^{topk} \in\mathbb{R}^{|T_{align}|\times k}$ and token sets $T_{stu}^{topk}, T_{tea}^{topk}$. At each token position $i\in [0,|T_{align}|)$, there are three possible situation for the tokens during mapping: 
\begin{itemize}
    \item The tokens can be exactly matched keep the 
 exact match mapping results(line 5)
    \item For tokens in $T^{topk}_{tea}[i]$ that cannot be exactly matched, search for the most similar token from the tokens in $T^{topk}_{stu}[i]$. To evaluate the similarity, we employ edit distance with length normalization with a similarity threshold $\theta$ to mitigate noisy matches (lines 6-15)
    \item For the tokens that cannot find a similar token to match, their corresponding logits will be masked before distillation. (lines 19-20)
\end{itemize}
During the training phase, $F_{dynamic}$ is continuously updated through this iterative process.
After the alignment of student output to teacher output, the vocabulary distribution of $O^{seq}_{stu}$ is projected via $F_{dynamic}$ to establish a shared probability space with $O^{seq}_{tea}$. To prevent noise from the mismatch vocabulary, we employ a masking operator $\text{Mask} (\cdot)$ that masks the logits for unmatched positions in the vocabulary dimension. This alignment mechanism yields the refined outputs $O_{tea}^{align}$ and $O_{stu}^{align}$.

To enhance the alignment of crucial tokens for student modeling, we implement a reverse-direction alignment from teacher to student. This process generates a reverse mapping dictionary $F^{reverse}_{dynamic}$ through analogous computational procedures. The alignment yields outputs: $O_{tea}^{reverse} = F^{reverse}_{dynamic}(O_{stu}^{topk})$ and $O_{stu}^{reverse}=\text{Mask}(O_{stu}^{topk})$. 
The final representations are constructed through vector concatenation (denoted by $\oplus$), formulated as: $O_{stu}^{f} = O_{stu}^{align}\oplus O_{stu}^{reverse}$ and $O_{tea}^{f} = O_{tea}^{align}\oplus O_{tea}^{reverse}$.
$O_{stu}^f \in \mathbb{R}^{|V_{align}|\times 2k}$ and $O_{tea}^f \in \mathbb{R}^{|V_{align}|\times 2k}$ have the same dimension and the same meaning correspond to concrete tokens or spans in sequence and vocabulary dimension.

\subsection{Aligned Logits Distillation}

After performing contextual alignment in both sequence and vocabulary dimensions, the distribution differences between the teacher and student logits are computed using the KL divergence, as shown in Equation~\ref{eq:kld}:
\begin{equation}\label{eq:kld} 
L_{\mathrm{KL}}(O_{stu}^{f} || O_{tea}^{f}) = \sum_{i=1}^k O_{stu}^{f}[i] \log \left( \frac{O_{stu}^{f}[i]}{O_{tea}^{f}[i]} \right) 
\end{equation}

Meanwhile, the language modeling target uses standard cross-entropy loss for a sentence $T=\{t_0, t_1, \dots \}$ and is defined as Equantion~\ref{eq:ce_loss}.
\begin{equation}\label{eq:ce_loss} L_{lm} = - \sum_{i}^{|T|} \log P(t_i | t_{i-1}, \cdots, t_1) \end{equation}
Integrated with the language modeling loss, weighted by $\alpha$, the final objective function is formulated in a manner analogous to classical model distillation, as presented in Equation~\ref{eq:final}.

\begin{equation}\label{eq:final} L = \alpha \cdot L_{\mathrm{KL}} + (1 - \alpha) \cdot L_{lm} \end{equation}

\section{Experiments}
In our experiments, we select five widely used open-source models with different architectures including Llama-3 (8B)~\cite{dubey2024llama}, OPT (1.3B/6.7B)~\cite{zhang2022opt}, Gemma-2 (2B/9B)~\cite{team2024gemma}, Phi3 (mini-3.8B)~\cite{abdin2024phi} and Qwen2 (0.5B/7B)\cite{yang2024qwen2}. As listed in Table~\ref{tab:tea_stu}, according to our statistic results in Section~\ref{sec:stat}, we select the combinations of the teacher model and student models from the pairs with relatively poor matching rates on vocabulary or sequence level.

\begin{table}[ht]
    \centering
    \resizebox{\linewidth}{!}{
    \begin{tabular}{llll}
    \toprule
         teacher & student & SMR & VMR \\
        \midrule
        Llama-3-8B & Gemma-2-2B & 85.52\% &67.79\%\\
         Llama-3-8B & OPT-1.3B &89.03\% &34.46\%\\
         Phi3-mini-3.8B& Qwen2-0.5B &31.54\%&65.65\% \\
         \bottomrule
    \end{tabular}}
    \caption{The settings on teacher and student models, where the SMR means the sequence matching rate and  VMR means the vocabulary matching rate.}
    \label{tab:tea_stu}
\end{table}

\begin{table}[ht]
    \centering
    \resizebox{\linewidth}{!}{
    \begin{tabular}{llcc}
    \toprule
    Type & Name & Number Train & Number Test \\
    \midrule
     \multirow{5}{*}{\makecell[l]{Instruction\\Following}} &Dolly   & 1100 & 500 \\
      & Self-Inst & - & 242 \\
      & Vicuna & - & 80 \\
      & S-NI & - & 1694 \\
      & UnNI & - & 1000 \\
      \hdashline
     \multirow{3}{*}{\makecell[l]{Code\\Generation}} & CodeM & 9600 & - \\
      & HumanEval+ & -& 164\\
      & MBPP & - & 500 \\
      \hdashline
      Math & GSM-8B & 7473 &1000 \\
      \bottomrule
    \end{tabular}
    }
    \caption{Data statistic of training and evaluation data.}
    \label{tab:data_stat}
\end{table}

\begin{table*}[ht]
    \centering
    \resizebox{\linewidth}{!}{
    \begin{tabular}{llccccc|c|cc|c}
    \toprule
     Type & Setting & Dolly & Self-Inst & Vicuna & S-NI & UnNI & \#Avg IF & HumanEval+ & MBPP+ & GSM-8K\\
    \midrule
        \multicolumn{11}{c}{\textit{Student Model: Gemma-2-2B}} \\
        \hdashline
     \multirow{3}{*}{SFT}   &Gemma-2-2B &25.12 & 14.94 & 16.89 & 25.29 & 30.07 & 22.46 &21.34&	21.34 &29.95 \\
        &Gemma-2-9B  &26.72 & 18.01 & 18.85 & 27.74 & 34.83 & 25.23&24.39 & 27.51 & 45.34 \\
        &Llama-3-8B &27.01 & 21.90 & 17.00 & 30.66 & 35.23 & 26.36 &34.76& 50.26  &44.20  \\
        \hdashline
       \multirow{2}{*}{\makecell[l]{Same Tokenizer KD\\\textit{(teacher: Gemma-2-9B)}}} &FKL  & 26.51 & 14.30 & 18.64 & 27.61 & 32.06 & 23.82 &18.90& 23.00 & 34.80  \\
        &RKL&25.26 & 13.80 & 18.64 & 23.70 & 29.79 & 22.24 & 18.90	&21.42&	27.37\\
        \hdashline
       \multirow{3}{*}{\makecell[l]{Cross Tokenizer KD\\\textit{(teacher: Llama-3-8B)}} }        &MinED &25.83	&\textbf{16.16}&	16.40&	25.99	&28.60&	22.60 &20.12	&\textbf{22.22}	&28.43 \\
        &ULD &26.11 & 14.58 & 17.25 & \textbf{27.69} & 30.53 & 23.23 & 20.40& 17.70  &26.38\\
       &CDM & \textbf{26.13}&	14.89&	\textbf{18.33}	&26.40	&\textbf{32.00}&	\textbf{23.55} & \textbf{23.78}& 21.69  &\textbf{30.40}\\
        \hline
        \multicolumn{11}{c}{\textit{Student Model: OPT-1.3B}} \\
        \hdashline
       \multirow{3}{*}{SFT}  &OPT-1.3B& 25.48&	14.26&	14.81&	25.88	&31.93&	22.47 & -- & -- & -- \\
        &OPT-6.7B& 28.40 & 15.71 & 15.82 & 26.87 & 33.56 & 24.07  & -- & -- & -- \\
        &Llama-3-8B &27.01 & 21.90 & 17.00 & 30.66 & 35.23 & 26.36  & -- & -- & -- \\
        \hdashline
      \multirow{2}{*}{\makecell[l]{Same Tokenizer KD\\(\textit{OPT-6.7B})}} &FKL &25.36 & 15.24 & 16.16 & 26.47 & 31.38 & 22.92 & -- & -- & -- \\
        &RKL& 25.03 & 13.24 & 15.42 & 23.86 & 31.27 & 21.77 & -- & -- & -- \\
        \hdashline
     \multirow{3}{*}{\makecell[l]{Cross Tokenizer KD\\\textit{(teacher: Llama-3-8B)}}}          &MinED & 25.21 & 12.60 & 15.60 & 24.51 & 30.52 & 21.69 & -- & -- & -- \\
        &ULD & 25.45 & 13.69 & \textbf{15.88} & 25.82 & 30.07 & 22.18 & -- & -- & -- \\
        &CDM &  \textbf{26.15} & \textbf{14.39} & 15.77 & \textbf{26.33} & \textbf{32.33} & \textbf{23.00} & -- & -- & -- \\
        \hline
        \multicolumn{11}{c}{\textit{Student Model: Qwen2-0.5B}} \\
        \hdashline
      \multirow{3}{*}{SFT}   &Qwen2-0.5B & 24.66 & 15.17 & 15.22 & 30.31 & 35.00 & 24.07 & 15.85&22.22&	27.22 \\
        &Qwen2-7B &29.07 & 22.69 & 21.42 & 37.31 & 41.04 & 30.31  & 39.02& 39.42   &59.14 \\
        &Phi3-mini-3.8B  & 29.19 & 25.39 & 21.81 & 37.97 & 41.07 & 31.09 & 51.83&	48.68&	64.67  \\
        \hdashline
      \multirow{2}{*}{\makecell[l]{Same Tokenizer KD\\(\textit{Qwen2-7B})}}  &FKL & 27.41 & 19.68 & 19.24 & 32.67 & 37.46 & 27.29 &17.07&	23.38&	27.67  \\
       & RKL & 26.15 & 16.15 &16.62&30.32&37.53 & 25.35& 20.73&	22.75&	26.38\\
         \hdashline
      \multirow{3}{*}{\makecell[l]{Cross Tokenizer KD\\\textit{(teacher: Phi3-mini-3.8B)}}}      & MinED & 25.55 & 16.26 & 15.37 & 30.76 & 35.69 & 24.72& 17.10&	22.20&	24.41\\
        &ULD & \textbf{26.43} & 16.15 & 15.34 & 30.63 & 36.07 & 24.93 &17.07&	22.49& 26.38\\
        &CDM &25.45 &\textbf{16.55} & \textbf{16.38} & \textbf{30.66} & \textbf{36.47} & \textbf{25.10} &\textbf{18.90}&	\textbf{23.81}&	\textbf{28.13} \\
        
    \bottomrule
    \end{tabular}
    }
    \caption{Main results of comparing CDM and the baseline models, where``\#AVG IF'' means the average score of the instruction-following tasks). The \textbf{blod} text means the best performance in comparable cross-tokenizer distillation settings. The table consists of three sections, each labeled with the student models in distillation experiments.}
    \label{tab:main_result}
\end{table*}

\subsection{Experiment Settings}
\paragraph{Baseline Methods}
In our experiments, the primary baseline constitutes supervised fine-tuning (SFT) applied to both teacher and student models.
To provide a comprehensive comparison, our baselines include the following methods for same-tokenizer model distillation (the teacher model maintains an identical architecture to the student model but with scaled-up parameters, {\em e.g,} Qwen2-7B serves as the teacher model for distilling knowledge into Qwen2-0.5B):
\begin{itemize}
    \item Forward KL divergence (FKL): the standard distillation loss function. Let $p(x)$ be the distribution of the student model, and $p(s)$ be the distribution of the teacher model, then the loss function is $L_{fkl}(p(x)||q(x))= \mathbb{E}_{q(x)}[log (\frac{q(x)}{p(x)}]$.
    \item Reverse KL divergence (RKL): reverse the distribution of teacher and student in KL divergence calculation.
\end{itemize}
And the following methods are for cross-tokenizer distillation:
\begin{itemize}
    \item Unified Logits Distance (ULD)~\cite{boizard2024towards}:  this cross-tokenizer distillation method leverages Optimal Transport to enable a unified distillation. 
    \item MinED~\cite{wan2024fusechat}: on the sequence level, the DTW uses edit distance as a cost function, while on the vocabulary level, it use exact match first and then uses edit distance to search the most similar token from the full vocabulary to get supplemental mappings.
\end{itemize}

\paragraph{Training Settings}
We use the learning rate 2e-5 and batch size 32 for Supervised Fine-Tuning (SFT) to train 10 epochs. For distillation methods, we follow the setting in MiniLLM~\cite{gu2024minillm}, the details of hyperparameters are appended in Section~\ref{sec:hyper}.
We evaluate the methods on three types of tasks: instruction-following, code generation, and math. The training datasets include Dolly-15K~\cite{ouyang2022training} for instruction following, CodeM~\cite{zan2024codem} for coding tasks, and GSM-8K for math tasks. The training is conducted on 8 Ascend 910Bs and using DeepSpeed\footnote{https://github.com/microsoft/DeepSpeed} ZeRO stage 1 for model parallel.

\paragraph{Evaluation Settings}
For instruction-following task, we follow the existing work~\cite{gu2024minillm, zhang2024dual} and evaluate Rouge-L~\cite{lin2004rouge} on a series of instruction-following test sets including Dolly , Self-Instruction~\cite{wang2023self} (Self-Inst), Vicuna-Evaluation~\cite{chiang2023vicuna} (Vicuna), Super-Natrul Instructions~\cite{wang2022benchmarking} (S-NI), and Unnatural Instruction~\cite{honovich2022unnatural} (UnNI). 
For code generation tasks, the test set contains HumanEval+~\cite{chen2021evaluating} and MBPP+~\cite{austin2021program} using EvalPlus~\cite{liu2024your} for an evaluation with more test cases and stricter.
The evaluate metric is Pass@1, meaning the ratio of generated code can pass all test cases in one shot, and the decoding setting is greedy search.
For math tasks, the test set is GSM-8K~\cite{cobbe2021training}. The evaluate metric is Test@1, which has the same meaning as Pass@1 for math problems, and the decoding setting is also greedy search. The data statistic of training and evaluation data can be referred to Table~\ref{tab:data_stat}.

\subsection{Main Results}
We conducted the fine-tuning and distillation experiments in the three settings of teacher-student. 
The main experiments of instruction-following are shown in Table~\ref{tab:main_result}, and there are two main findings.
First, among all settings, the performance of CDM is significantly higher than other cross-tokenizer distillation baselines ({\em e.g.}, CDM outperforms ULD by around 0.88 average Rouge-L). Second, compared with the same-tokenizer distillation methods, including FKL and RKL, CDM achieves better performance in OPT and significantly exceeds the related method of cross-tokenizer distillation.

For code generation and mathematical reasoning tasks, we excluded OPT models due to their inadequate pre-training in these specialized domains, which fundamentally limits knowledge distillation efficacy. 
Both Gemma and Qwen models achieved varying degrees of performance gains through distillation, with the CDM method consistently demonstrating the most stable and superior effectiveness among cross-tokenizer approaches. Particularly, Qwen2-0.5B delivers notable average improvements of 12.19\% on code generation and 3.34\% on math tasks. 
 These consistent improvements across three task categories substantiate CDM's effectiveness, with comprehensive analyses provided in Section~\ref{sec:alignment}.

\section{Analysis}
Our analysis experiments adopts Phi3-mini-3.8B and Qwen2-0.5B as a default configuration, mainly including quantitative measurements of sequence-level and vocabulary-level alignment improvements (Section~\ref{sec:alignment}), comparative analysis with sequence-level knowledge distillation (Section~\ref{sec:seqKD}), and systematic exploration of dual-teacher distillation paradigms (Section~\ref{sec:dual_tea}). We introduce "Average IF" as a composite metric aggregating performance across five instruction-following tasks to streamline result interpretation. Moreover, the supplementary analysis, including ablation studies (Section~\ref{sec:ablation}), sensitivity analyses (Section~\ref{sec:sensitivity}), and comparative case studies of alignment outcomes (Sections~\ref{sec:case_seq}-\ref{sec:case_tok}), are comprehensively documented in the Appendix.

\subsection{The Statistic on Alignment Rate}\label{sec:alignment}
In this section, we analyze the extent of the sequence-level and vocabulary-level alignment improvement separately.
\paragraph{Sequence Level}
We define the sequence alignment rate through the following procedure. When aligning two sequences, adjacent tokens may merge into contiguous spans for correspondence mapping.
Similar to the statistic in Section~\ref{sec:stat}, the correspondent span pairs that cannot exactly match will be regarded as a mismatch.
We sampled 3,000 sentences from the training set of Dolly. The alignment rate of the results in the pure edit distance cost function was 78.31\%, which improved to 82.20\% after adding entropy weight, demonstrating our method's efficacy. Detailed case studies supporting these findings are presented in Section\ref{sec:case_seq}.

\paragraph{Vocabulary Level}
On the vocabulary level, CDM optimizes noise suppression and coverage in vocabulary mapping at the same time.
Exact Match (EM) provides unambiguous alignment but leads to limited mapping coverage, while similarity-based fuzzy matching inevitably introduces erroneous mappings that negatively impact model distillation effectiveness. 
On the mapping coverage, the basic EM between Qwen2 and Phi3 achieves 65\% on the coverage rate, and CDM significantly improves to 87\%.
There are two representative offline vocabulary matching approaches, including mapping using edit distance (ED)~\cite{wan2024fusechat} and edit distance with prefix constrain(PrefixED)~\cite{wu2024cross}, and both of them achieve over 99\% coverage rates. 
To ensure rigorous comparison with existing approaches, we conduct controlled experiments maintaining identical experimental conditions except for the vocabulary mapping strategy. As shown in Table~\ref{tab:mapping_comp}, outperforms these high-coverage methods by substantial margins, confirming its superior noise suppression capability. Extended case studies with detailed pattern analyses are available in Appendixn~\ref{sec:case_tok}.
\begin{table}[ht]
    \centering
    \resizebox{0.44\linewidth}{!}{
    \begin{tabular}{lc}
    \toprule
        Setting &   Average IF \\
        \midrule
        ED &  24.32\\
        PrefixED & 24.12\\
        CDM  & \textbf{25.10}\\
        \bottomrule
    \end{tabular}
    }
    \caption{Comparison between different vocabulary mapping methods.}
    \label{tab:mapping_comp}
\end{table}

\subsection{Comparison with Sequence-Level KD}~\label{sec:seqKD}
In this section, we provide a comparison between our method and sequence-level KD~\cite{kim2016sequence} (SeqKD).
Unlike the probability-based methods discussed in the main experiments, SeqKD uses the generated text of the teacher model to enhance the student model performance. In the most advanced models, SeqKD is also applied for cross-tokenizer distillation seneraios~\cite{guo2025deepseek}.\footnote{Our implementation employs sampling decoding (Temperature=0.2) and integrates teacher-generated data with original training data, conducting 5-epoch training to maintain equivalent training iterations.}
According to Table~\ref{tab:seqkd}, the results demonstrate that SeqKD and SFT have close performance, which indicates the necessity of logit-based distillation.
\begin{table}[ht]
    \centering
    \resizebox{0.42\linewidth}{!}{
    \begin{tabular}{lc}
    \toprule
      Setting & Average IF \\
    \midrule
    SFT  & 24.07  \\
        SeqKD & 24.05 \\
        CDM  & \textbf{25.10}\\
    \bottomrule
    \end{tabular}
    }
    \caption{The comparison between CDM and SeqKD.}
    \label{tab:seqkd}
\end{table}

\subsection{Dual-teacher Distillation}\label{sec:dual_tea}

To investigate the impact of knowledge distillation across different model architectures, we adopted an integrated approach that combines both same-tokenizer and cross-tokenizer knowledge distillation based on OPT-1.3B. Although the results indicate limited improvement over SFT for both FKD and CDM, a significant improvement (10.46\% in instruction-following tasks) was observed when simultaneously leveraging the knowledge of two distinct teacher models through an average loss of distillation.
The findings in dual-teacher settings imply that different tokenizer strategies may provide complementary information that is effectively utilized to enhance model performance.

\begin{table}[ht]
    \centering
    \resizebox{0.65\linewidth}{!}{
    \begin{tabular}{lc}
    \toprule
    Settings & Average IF \\
    \midrule
       SFT  & 22.47 \\ 
       w/ OPT-6.7B (FKD) & 22.92\\
       w/ Llama-3-8B (CDM) & 23.00 \\
       Dual Teacher & \textbf{24.82}\\
         \bottomrule
    \end{tabular}}
    \caption{The comparison for distillation using dual-teacher and single-teacher settings.}
    \label{tab:dual_tea}
\end{table}

\section{Related Work}
In the context of cross-tokenizer distillation, several methods have been proposed to align the probability distributions of models before performing distillation. This alignment typically involves both sequence and vocabulary dimensions.
~\citet{fu2023specializing} aligns sequences through dynamic programming and aligns vocabularies through exact matching. To improve vocabulary alignment, the~\cite{wanknowledge,wan2024fusechat} series introduced methods such as MinED and statistical matching for fuzzy matching to supplement vocabulary alignment.
Despite these advancements, their effectiveness remains constrained by the prevalence of numerous mismatches. Beyond alignment strategies based on text character similarity, \citet{boizard2024towards} proposed the use of optimal transport to quantify the distance between model logits. Furthermore, \citet{cui2024multi} refine this approach by optimizing the cost function at both sequence and vocabulary levels, effectively integrating both local and global information to improve overall performance. Additionally, DSKD~\cite{zhang2024dual} introduces a dual alignment framework that simultaneously aligns hidden states and model logits. However, both ULD and DSKD methods suffer from inefficiencies due to underutilization of the vocabulary. 

\section{Conclusion and Future Work}

In this work, we propose Contextual Dynamic Mapping (CDM), a novel approach to CKTD. CDM enhances the CTKD by improving the alignment of model outputs on both sequence and vocabulary levels through the use of online context information. Specifically, the method incorporates entropy-based weights in the sequential alignment process and employs contextual Top-K token pairs to dynamically map vocabulary probabilities.
Through extensive experiments across three groups of teacher-student configurations and three types of tasks (instruction-following, code generation and math), our method demonstrates significant advantages over existing approaches and shows further potential in dual-teacher scenarios. Furthermore, statistical analyses and case studies are presented to demonstrate how the method enhances model alignment accuracy in both sequence and vocabulary.
In future work, we plan to further scale the application of CDM by assessing its performance on more diverse training datasets and larger 
student and teacher models to enhance scalability. In particular, the dual-teacher distillation will be extended to multi-teacher settings for further observation.

\bibliography{custom}

\begin{thebibliography}{29}
\providecommand{\natexlab}[1]{#1}

\bibitem[{Abdin et~al.(2024)Abdin, Aneja, Awadalla, Awadallah, Awan, Bach, Bahree, Bakhtiari, Bao, Behl et~al.}]{abdin2024phi}
Marah Abdin, Jyoti Aneja, Hany Awadalla, Ahmed Awadallah, Ammar~Ahmad Awan, Nguyen Bach, Amit Bahree, Arash Bakhtiari, Jianmin Bao, Harkirat Behl, et~al. 2024.
\newblock Phi-3 technical report: A highly capable language model locally on your phone.
\newblock \emph{arXiv preprint arXiv:2404.14219}.

\bibitem[{Austin et~al.(2021)Austin, Odena, Nye, Bosma, Michalewski, Dohan, Jiang, Cai, Terry, Le et~al.}]{austin2021program}
Jacob Austin, Augustus Odena, Maxwell Nye, Maarten Bosma, Henryk Michalewski, David Dohan, Ellen Jiang, Carrie Cai, Michael Terry, Quoc Le, et~al. 2021.
\newblock Program synthesis with large language models.
\newblock \emph{arXiv preprint arXiv:2108.07732}.

\bibitem[{Boizard et~al.(2024)Boizard, Haddad, Hudelot, and Colombo}]{boizard2024towards}
Nicolas Boizard, Kevin~El Haddad, C{\'e}line Hudelot, and Pierre Colombo. 2024.
\newblock Towards cross-tokenizer distillation: the universal logit distillation loss for llms.
\newblock \emph{arXiv preprint arXiv:2402.12030}.

\bibitem[{Chen et~al.(2021)Chen, Tworek, Jun, Yuan, Pinto, Kaplan, Edwards, Burda, Joseph, Brockman et~al.}]{chen2021evaluating}
Mark Chen, Jerry Tworek, Heewoo Jun, Qiming Yuan, Henrique Ponde De~Oliveira Pinto, Jared Kaplan, Harri Edwards, Yuri Burda, Nicholas Joseph, Greg Brockman, et~al. 2021.
\newblock Evaluating large language models trained on code.
\newblock \emph{arXiv preprint arXiv:2107.03374}.

\bibitem[{Chiang et~al.(2023)Chiang, Li, Lin, Sheng, Wu, Zhang, Zheng, Zhuang, Zhuang, Gonzalez et~al.}]{chiang2023vicuna}
Wei-Lin Chiang, Zhuohan Li, Zi~Lin, Ying Sheng, Zhanghao Wu, Hao Zhang, Lianmin Zheng, Siyuan Zhuang, Yonghao Zhuang, Joseph~E Gonzalez, et~al. 2023.
\newblock Vicuna: An open-source chatbot impressing gpt-4 with 90\%* chatgpt quality.
\newblock \emph{See https://vicuna. lmsys. org (accessed 14 April 2023)}, 2(3):6.

\bibitem[{Cobbe et~al.(2021)Cobbe, Kosaraju, Bavarian, Chen, Jun, Kaiser, Plappert, Tworek, Hilton, Nakano et~al.}]{cobbe2021training}
Karl Cobbe, Vineet Kosaraju, Mohammad Bavarian, Mark Chen, Heewoo Jun, Lukasz Kaiser, Matthias Plappert, Jerry Tworek, Jacob Hilton, Reiichiro Nakano, et~al. 2021.
\newblock Training verifiers to solve math word problems.
\newblock \emph{arXiv preprint arXiv:2110.14168}.

\bibitem[{Cui et~al.(2024)Cui, Zhu, Qin, Xie, Zhou, and Li}]{cui2024multi}
Xiao Cui, Mo~Zhu, Yulei Qin, Liang Xie, Wengang Zhou, and Houqiang Li. 2024.
\newblock Multi-level optimal transport for universal cross-tokenizer knowledge distillation on language models.
\newblock \emph{arXiv preprint arXiv:2412.14528}.

\bibitem[{Dubey et~al.(2024)Dubey, Jauhri, Pandey, Kadian, Al-Dahle, Letman, Mathur, Schelten, Yang, Fan et~al.}]{dubey2024llama}
Abhimanyu Dubey, Abhinav Jauhri, Abhinav Pandey, Abhishek Kadian, Ahmad Al-Dahle, Aiesha Letman, Akhil Mathur, Alan Schelten, Amy Yang, Angela Fan, et~al. 2024.
\newblock The llama 3 herd of models.
\newblock \emph{arXiv preprint arXiv:2407.21783}.

\bibitem[{Fu et~al.(2023)Fu, Peng, Ou, Sabharwal, and Khot}]{fu2023specializing}
Yao Fu, Hao Peng, Litu Ou, Ashish Sabharwal, and Tushar Khot. 2023.
\newblock Specializing smaller language models towards multi-step reasoning.
\newblock In \emph{International Conference on Machine Learning}, pages 10421--10430. PMLR.

\bibitem[{Gu et~al.(2024)Gu, Dong, Wei, and Huang}]{gu2024minillm}
Yuxian Gu, Li~Dong, Furu Wei, and Minlie Huang. 2024.
\newblock Minillm: Knowledge distillation of large language models.
\newblock In \emph{The Twelfth International Conference on Learning Representations}.

\bibitem[{Guo et~al.(2025)Guo, Yang, Zhang, Song, Zhang, Xu, Zhu, Ma, Wang, Bi et~al.}]{guo2025deepseek}
Daya Guo, Dejian Yang, Haowei Zhang, Junxiao Song, Ruoyu Zhang, Runxin Xu, Qihao Zhu, Shirong Ma, Peiyi Wang, Xiao Bi, et~al. 2025.
\newblock Deepseek-r1: Incentivizing reasoning capability in llms via reinforcement learning.
\newblock \emph{arXiv preprint arXiv:2501.12948}.

\bibitem[{Hinton(2015)}]{hinton2015distilling}
Geoffrey Hinton. 2015.
\newblock Distilling the knowledge in a neural network.
\newblock \emph{arXiv preprint arXiv:1503.02531}.

\bibitem[{Honovich et~al.(2022)Honovich, Scialom, Levy, and Schick}]{honovich2022unnatural}
Or~Honovich, Thomas Scialom, Omer Levy, and Timo Schick. 2022.
\newblock Unnatural instructions: Tuning language models with (almost) no human labor.
\newblock \emph{arXiv preprint arXiv:2212.09689}.

\bibitem[{Kim and Rush(2016)}]{kim2016sequence}
Yoon Kim and Alexander~M Rush. 2016.
\newblock Sequence-level knowledge distillation.
\newblock In \emph{Proceedings of the 2016 Conference on Empirical Methods in Natural Language Processing}, pages 1317--1327.

\bibitem[{Ko et~al.(2024)Ko, Kim, Chen, and Yun}]{ko2024distillm}
Jongwoo Ko, Sungnyun Kim, Tianyi Chen, and Se-Young Yun. 2024.
\newblock Distillm: Towards streamlined distillation for large language models.
\newblock \emph{arXiv preprint arXiv:2402.03898}.

\bibitem[{Lin(2004)}]{lin2004rouge}
Chin-Yew Lin. 2004.
\newblock Rouge: A package for automatic evaluation of summaries.
\newblock In \emph{Text summarization branches out}, pages 74--81.

\bibitem[{Liu et~al.(2024)Liu, Xia, Wang, and Zhang}]{liu2024your}
Jiawei Liu, Chunqiu~Steven Xia, Yuyao Wang, and Lingming Zhang. 2024.
\newblock Is your code generated by chatgpt really correct? rigorous evaluation of large language models for code generation.
\newblock \emph{Advances in Neural Information Processing Systems}, 36.

\bibitem[{Ouyang et~al.(2022)Ouyang, Wu, Jiang, Almeida, Wainwright, Mishkin, Zhang, Agarwal, Slama, Ray et~al.}]{ouyang2022training}
Long Ouyang, Jeffrey Wu, Xu~Jiang, Diogo Almeida, Carroll Wainwright, Pamela Mishkin, Chong Zhang, Sandhini Agarwal, Katarina Slama, Alex Ray, et~al. 2022.
\newblock Training language models to follow instructions with human feedback.
\newblock \emph{Advances in neural information processing systems}, 35:27730--27744.

\bibitem[{Team et~al.(2024)Team, Riviere, Pathak, Sessa, Hardin, Bhupatiraju, Hussenot, Mesnard, Shahriari, Ram{\'e} et~al.}]{team2024gemma}
Gemma Team, Morgane Riviere, Shreya Pathak, Pier~Giuseppe Sessa, Cassidy Hardin, Surya Bhupatiraju, L{\'e}onard Hussenot, Thomas Mesnard, Bobak Shahriari, Alexandre Ram{\'e}, et~al. 2024.
\newblock Gemma 2: Improving open language models at a practical size.
\newblock \emph{arXiv preprint arXiv:2408.00118}.

\bibitem[{Wan et~al.(2024{\natexlab{a}})Wan, Huang, Cai, Quan, Bi, and Shi}]{wanknowledge}
Fanqi Wan, Xinting Huang, Deng Cai, Xiaojun Quan, Wei Bi, and Shuming Shi. 2024{\natexlab{a}}.
\newblock Knowledge fusion of large language models.
\newblock In \emph{The Twelfth International Conference on Learning Representations}.

\bibitem[{Wan et~al.(2024{\natexlab{b}})Wan, Zhong, Yang, Chen, and Quan}]{wan2024fusechat}
Fanqi Wan, Longguang Zhong, Ziyi Yang, Ruijun Chen, and Xiaojun Quan. 2024{\natexlab{b}}.
\newblock Fusechat: Knowledge fusion of chat models.
\newblock \emph{arXiv preprint arXiv:2408.07990}.

\bibitem[{Wang et~al.(2023)Wang, Kordi, Mishra, Liu, Smith, Khashabi, and Hajishirzi}]{wang2023self}
Yizhong Wang, Yeganeh Kordi, Swaroop Mishra, Alisa Liu, Noah~A Smith, Daniel Khashabi, and Hannaneh Hajishirzi. 2023.
\newblock Self-instruct: Aligning language models with self-generated instructions.
\newblock In \emph{Proceedings of the 61st Annual Meeting of the Association for Computational Linguistics (Volume 1: Long Papers)}, pages 13484--13508.

\bibitem[{Wang et~al.(2022)Wang, Mishra, Alipoormolabashi, Kordi, Mirzaei, Arunkumar, Ashok, Dhanasekaran, Naik, Stap et~al.}]{wang2022benchmarking}
Yizhong Wang, Swaroop Mishra, Pegah Alipoormolabashi, Yeganeh Kordi, Amirreza Mirzaei, Anjana Arunkumar, Arjun Ashok, Arut~Selvan Dhanasekaran, Atharva Naik, David Stap, et~al. 2022.
\newblock Benchmarking generalization via in-context instructions on 1,600+ language tasks.
\newblock \emph{arXiv preprint arXiv:2204.07705}, 2.

\bibitem[{Wen et~al.(2023)Wen, Li, Du, and Mou}]{wen2023f}
Yuqiao Wen, Zichao Li, Wenyu Du, and Lili Mou. 2023.
\newblock f-divergence minimization for sequence-level knowledge distillation.
\newblock In \emph{Proceedings of the 61st Annual Meeting of the Association for Computational Linguistics (Volume 1: Long Papers)}, pages 10817--10834.

\bibitem[{Wu et~al.(2024)Wu, Sun, Cai, Su, Wang, Yin, Li, and Gao}]{wu2024cross}
Jiayi Wu, Hao Sun, Hengyi Cai, Lixin Su, Shuaiqiang Wang, Dawei Yin, Xiang Li, and Ming Gao. 2024.
\newblock Cross-model control: Improving multiple large language models in one-time training.
\newblock In \emph{The Thirty-eighth Annual Conference on Neural Information Processing Systems}.

\bibitem[{Yang et~al.(2024)Yang, Yang, Hui, Zheng, Yu, Zhou, Li, Li, Liu, Huang et~al.}]{yang2024qwen2}
An~Yang, Baosong Yang, Binyuan Hui, Bo~Zheng, Bowen Yu, Chang Zhou, Chengpeng Li, Chengyuan Li, Dayiheng Liu, Fei Huang, et~al. 2024.
\newblock Qwen2 technical report.
\newblock \emph{arXiv preprint arXiv:2407.10671}.

\bibitem[{Zan et~al.(2024)Zan, Yu, Liu, Shen, Lin, Gong, Yao, Liu, Guan, Luo et~al.}]{zan2024codem}
Daoguang Zan, Ailun Yu, Wei Liu, Bo~Shen, Shaoxin Lin, Yongshun Gong, Yafen Yao, Yan Liu, Bei Guan, Weihua Luo, et~al. 2024.
\newblock Codem: Less data yields more versatility via ability matrix.
\newblock In \emph{Findings of the Association for Computational Linguistics ACL 2024}, pages 714--729.

\bibitem[{Zhang et~al.(2024)Zhang, Zhang, Sun, Chen, and Xu}]{zhang2024dual}
Songming Zhang, Xue Zhang, Zengkui Sun, Yufeng Chen, and Jinan Xu. 2024.
\newblock Dual-space knowledge distillation for large language models.
\newblock In \emph{Proceedings of the 2024 Conference on Empirical Methods in Natural Language Processing}, pages 18164--18181.

\bibitem[{Zhang et~al.(2022)Zhang, Roller, Goyal, Artetxe, Chen, Chen, Dewan, Diab, Li, Lin et~al.}]{zhang2022opt}
Susan Zhang, Stephen Roller, Naman Goyal, Mikel Artetxe, Moya Chen, Shuohui Chen, Christopher Dewan, Mona Diab, Xian Li, Xi~Victoria Lin, et~al. 2022.
\newblock Opt: Open pre-trained transformer language models.
\newblock \emph{arXiv preprint arXiv:2205.01068}.

\end{thebibliography}

\appendix
\section{Appendix}
\subsection{Details of Training Settings in Main Experiemnts}\label{sec:hyper}
 For all SFT baselines, the last checkpoint of 10 epochs is applied to be the final result. For distillation settings, three-epoch SFT is conducted first, and then seven epochs of continual distillation are applied to keep the total steps the same. The hyper-parameters about distillation are as Table~\ref{tab:hyper-param}.
\begin{table}[ht]
    \centering
    \resizebox{0.4\linewidth}{!}{
    \begin{tabular}{lc}
    \toprule
        Notation & Value \\
        \midrule
       $\theta$  & 0.3 \\
       $K$ & 100\\
       $\alpha$ & 0.5 \\
       $T$ & 2.0 \\
       $C$ & 3 \\
       \bottomrule
    \end{tabular}
    }
    \caption{Hyper-paramenters of the distillation methods.}
    \label{tab:hyper-param}
\end{table}

\subsection{Ablation Study}\label{sec:ablation}
To verify the effectiveness of the modules of the method independently, we conducted ablation studies comparing three configurations: (1) removing sequence-level dynamic alignment, (2) using exact match directly and disabling dynamic vocabulary matching and (3) eliminating the daul mapping strategy  ({\em i.e.}, cancel the reverse process in Section~\ref{sec:cdm_voc}, and only calculate distillation loss based on the Top-K logits of the teacher model and its correspondent logits in the student model outputs).
As evidenced by Table~\ref{tab:ablation}, each component removal adversely affects model convergence. Specifically, disabling sequence-level alignment reduces performance by 1.23 points on average, while removing vocabulary-level matching and dual mapping cause 0.86-point and 1.25-point degradations, respectively. These results quantitatively demonstrate the necessity of simultaneous optimization across both sequence and vocabulary dimensions and dual mapping.

\begin{table}[ht]
    \centering
    \resizebox{\linewidth}{!}{
    \begin{tabular}{lc}
    \toprule
      Setting & Average IF \\
    \midrule
    CDM  & \textbf{25.10}\\
    
        - w/o Entropy-based Weight   & 23.87\\
        - w/o Dynamic Vocabuaray Mapper & 24.24 \\
        - w/o Dual Mapping & 23.85 \\
        SFT  & 24.07  \\
        
    \bottomrule
    \end{tabular}
    }
    \caption{Ablation study on CDM.}
    \label{tab:ablation}
\end{table}

\subsection{On the Sensitivity of Hyper-Parameters}~\label{sec:sensitivity}
This section systematically examines the parameter sensitivity of two critical hyperparameters in our CDM framework: the similarity threshold ($\theta$) and $k$ in the Top-K selection.
\paragraph{On Similarity Threshold}
The similarity threshold $\theta$ is the key to controlling the selection of candidate tokens for mapping, which can be referred to in Algorithm~\ref{alg:AOS}.
In this section, we conduct a comprehensive sensitivity analysis by evaluating four representative similarity thresholds: 0.0, 0.1, 0.3, and 0.5, as detailed in Table~\ref{tab:simi}. To ensure consistent evaluation across tokens of varying lengths, we employed an edit-distance score normalized by token length. This approach ensures that a threshold of 0.0 corresponds to exact string matching, while a threshold of 0.5 permits a broader range of fuzzy matches, albeit at the expense of precision. Our experimental results demonstrate that an intermediate threshold strikes a better balance between accuracy and coverage, with thresholds of 0.1 and 0.3 emerging as particularly effective in this context.

\begin{table}[ht]
    \centering
    \resizebox{0.4\linewidth}{!}{
    \begin{tabular}{lc}
    \toprule
      $\theta$ & Average IF \\
    \midrule
        0.0 & 24.24 \\
        0.1 & 24.98 \\
        0.3 & \textbf{25.10} \\
        0.5 & 24.31 \\
    \bottomrule
    \end{tabular}
    }
    \caption{Sensitivity test on hyperparameter similarity threshold, where 0.0 means exact matching, and 0.5 means a relatively low requirement on similarity.}
    \label{tab:simi}
\end{table}

\paragraph{On the selection of K for Top-K}
The determination of $k$ in Top-K sampling constitutes a pivotal design consideration for Cross-Domain Mapping (CDM), as it directly influences the number and diversity of candidate tokens during vocabulary mapping.
To systematically evaluate how this hyperparameter affects model convergence, we conducted controlled experiments across multiple $k$ values. Empirical results demonstrate that while all Top-K configurations outperform supervised fine-tuning baselines, optimal model performance occurs at $k=100$. Conversely,  performance degradation emerges at both extremes: undersized candidate pools (K=50) restrict mapping flexibility through excessive token exclusion, while oversized pools (K=200) introduce noise token mappings. This non-monotonic relationship is quantitatively validated in Table~\ref{tab:topk}.

\begin{table}[ht]
    \centering
    \resizebox{0.45\linewidth}{!}{
    \begin{tabular}{lc}
    \toprule
     $k$ & Average IF \\
    \midrule
        50 & 24.37 \\
        100 & \textbf{25.10} \\
        200 & 24.70 \\
        500 & 24.66 \\
    \bottomrule
    \end{tabular}
    }
    \caption{Sensitivty test on hyperparameter Top-K}
    \label{tab:topk}
\end{table}

\subsection{Case Study on Sequence Alignment}\label{sec:case_seq}
The following cases in Table~\ref{tab:case_seq} demonstrate the effectiveness of the entropy-based weight in enhancing the DTW alignment algorithm. In the original DTW alignment results, misalignments occur frequently at positions with ambiguous meanings, such as commas or short alphabet spans at the beginning of sentences. In contrast, the entropy-weighted DTW approach in our method (CDM) achieves more accurate span alignment. By assigning higher weights to these ambiguous positions, which carry more information, the underestimation of their importance is reduced, resulting in improved alignment accuracy. This case study demonstrates that context-aware weight calibration substantially improves alignment robustness for linguistically ambiguous elements.

\begin{table*}[h]
    \centering
    \begin{tabular}{l|l|l}
    \toprule
    id & type        & content       \\
    \midrule
    1 & A & Moon |||  Knight |||  is |||  Marvel ||| , |||  Batman |||  is |||  DC  \\
    & B           & Moon ||| Knight ||| is ||| Marvel ||| , ||| Bat ||| man ||| is ||| DC \\
    \hdashline
    & A'        & Moon ||| Knight ||| is ||| Marvel |||{\color{red} , ||| Batman }||| is ||| DC    \\
    & B'          & Moon ||| Knight ||| is ||| Marvel |||{\color{red} ,Bat ||| man }||| is ||| DC \\
    \hdashline
    & CDM A' & Moon ||| Knight ||| is ||| Marvel ||| , ||| Batman ||| is ||| DC \\
    & CDM B'  & Moon ||| Knight ||| is ||| Marvel ||| , ||| Batman ||| is ||| DC \\
                    \hline
    2     & A  & Ant ||| -Man |||  is |||  Marvel ||| , |||  Ray |||  Palmer |||  is |||  DC   \\
        & B           & Ant ||| - ||| Man ||| is ||| Marvel ||| , ||| Ray ||| Pal ||| mer ||| is ||| DC \\
        \hdashline
        & A'          & {\color{red}Ant ||| -Man }||| is ||| Marvel ||| , |||{\color{red} Ray ||| Palmer }||| is ||| DC\\
        & B'          & {\color{red}Ant- ||| Man }||| is ||| Marvel ||| , |||{\color{red} RayPal ||| mer }||| is ||| DC \\
        \hdashline
        & CDM A' & {\color{red}Ant ||| -Man }||| is ||| Marvel ||| , ||| Ray ||| Palmer ||| is ||| DC \\
        & CDM B'  & {\color{red}Ant- ||| Man} ||| is ||| Marvel ||| , ||| Ray ||| Palmer ||| is ||| DC \\
                       \hline
    3     & A           & D ||| odge |||  is |||  American ||| , |||  Volkswagen |||  is |||  German      \\
                       & B           & D ||| odge ||| is ||| American ||| , ||| Volks ||| wagen ||| is ||| German      \\
                       \hdashline
                       & A'          & D ||| odge ||| is ||| American |||{\color{red} , ||| Volkswagen} ||| is ||| German           \\
                       & B'          & D ||| odge ||| is ||| American |||{\color{red} ,Volks ||| wagen }||| is ||| German           \\
                       \hdashline
                       & CDM A' & D ||| odge ||| is ||| American ||| , ||| Volkswagen ||| is ||| German           \\
                       & CDM B'  & D ||| odge ||| is ||| American ||| , ||| Volkswagen ||| is ||| German   \\    
                       \bottomrule
    \end{tabular}
    \caption{Three representative examples of sequence alignment outcomes demonstrate the impact of incorporating contextual information. Noting that although CDM is also not fully correct on the second case, its misalignment parts do not affect the main token meaning.}
    \label{tab:case_seq}
\end{table*}

\subsection{Case Study on Vocabulary Alignment}\label{sec:case_tok}
Table~\ref{tab:case_tok} provides case studies demonstrating how contextual information optimizes vocabulary mapping accuracy in CTKD. Our contextual alignment mechanism successfully resolves lexical ambiguities by integrating character-similarity and semantic dependencies. In contrast, conventional edit-distance approaches exhibit fundamental limitations: (1) inability to capture semantic relationships beyond surface-form similarity, and (2) lack of dynamic adaptation to contextual variations. The contrast reveals that context-agnostic methods relying solely on character-level edit operations systematically neglect higher-order semantic associations crucial for knowledge transfer.
\begin{table*}[h]
    \centering
    \begin{tabular}{l|l|p{8cm}}
    \toprule
    id & type          & content  \\
    \midrule
    1 & full sentence & There isn't any one bicycle that would be ideal for all people. Bike shops have experts who can advise the right model and size for you and your main uses. You could also look at product results from online bike shops and read reviews to supplement the advice from the shop. A meetup or ride with a local cycling group would be another great source of advice and targeted knowledge for making a decision which bike is right for you. \\
    \hdashline
     & w/o context &"\_publishers"$\rightarrow$"gepubliceerd"\\
     \hdashline
     & w/ context& "\_publishers"$\rightarrow$ $\emptyset$\\ 
     \hline
    2     & full sentence & The English army fought for King Harold Godwinson. \\
    \hdashline
& w/o context   & "\_fights"$\rightarrow$"weights"   \\
\hdashline
 & w/ context    & "\_fights$\rightarrow$"fight"  \\
\hline
3     & full sentence & Championship rowing races are conducted over 2 kilometers (1.2 miles) with dedicated lanes delineated by bouys.    \\
\hdashline
 & w/o context & "kilom" $\rightarrow$ ".iloc";"denoted"$\rightarrow$"\_devoted" \\
\hdashline
& w/ context    & "kilom"$\rightarrow$ " kilomet"; "denoted"$\rightarrow$"\_defined" \\
                           \bottomrule
    \end{tabular}
    \caption{Three representative examples of token mapping are presented, where the prefix `\_' denotes a space. The tokens are from the Top-K tokens, so they may not appear in the original full text.}
    \label{tab:case_tok}
\end{table*}

\end{document}